\documentclass[conference]{IEEEtran}
\IEEEoverridecommandlockouts
\usepackage{cite}
\usepackage{amsmath,amssymb,amsfonts}
\usepackage{algorithmic}
\usepackage{graphicx}
\usepackage{multirow}
\usepackage{textcomp}
\usepackage{xcolor}
\def\BibTeX{{\rm B\kern-.05em{\sc i\kern-.025em b}\kern-.08em
    T\kern-.1667em\lower.7ex\hbox{E}\kern-.125emX}}
\begin{document}

\title{InDEX: Indonesian Idiom and Expression Dataset for Cloze Test\\
\thanks{This paper is partially supported by Guangzhou Science and Technology Plan Project 202201010729 and GKMLP Grant No. GKLMLP-2021-001.}
}

\author{\IEEEauthorblockN{Xinying Qiu}
\IEEEauthorblockA{\textit{Guangzhou Key Laboratory of Multilingual Intelligent Processing} \\
\textit{Guangdong University of Foreign Studies}\\
Guangzhou, China \\
xy.qiu@foxmail.com}
\and
\IEEEauthorblockN{Guofeng Shi}
\IEEEauthorblockA{\textit{School of Information Science and Technology} \\
\textit{Guangdong University of Foreign Studies}\\
Guangzhou, China \\
s\_guofeng@foxmail.com}
}

\maketitle

\begin{abstract}
We propose InDEX, an Indonesian Idiom and Expression dataset for cloze test.  The dataset contains 10438 unique sentences for 289 idioms and expressions for which we generate 15 different types of distractors, resulting in a large cloze-style corpus. Many baseline models of cloze test reading comprehension apply BERT with random initialization to learn embedding representation. But idioms and fixed expressions are different such that the literal meaning of the phrases may or may not be consistent with their contextual meaning. Therefore, we explore different ways to combine static and contextual representations for a stronger baseline model. Experimentations show that combining definition and random initialization will better support cloze test model performance for idioms whether independently or mixed with fixed expressions. While for fixed expressions with no special meaning, static embedding with random initialization is sufficient for cloze test model.
\end{abstract}


\section{Introduction}
Cloze-style test is one of the machine reading comprehension tasks. The main goal is to answer a multiple-choice question where a key word is left blank in the contextual sentence or paragraph. Corpus construction for idiom reading comprehension has been studied for many different languages such as Japanese~\cite{b1}, English(\cite{b2, b3}), Portugese~\cite{b4}, and Chinese(\cite{b5,b6}), but very rare for non-universal langauges such as Indonesian. In addition, most of the available datasets and reading comprehension models for cloze test focus on idioms only, without considering or comparisons with fixed expressions. 
     
     A fixed expression is a fully lexicalized combination of words which are not flexible in patterning but transparent in meaning in general. An idiom, however, has a separate meaning of its own that are completely different from the lexcial meaning of its compositional words.  As shown in Table~\ref{table1}, the individual words of a fixed expression still maintain a lexical connection. But the meaning of idioms can not be deduced from its individual words. In language learning, fixed expressions and idioms are important knowledge especially in advanced study. For example, in Cambridge Advance Exam, test of understading fixed expression and idioms in context plays a huge part and poses a much greater challenge to learners at advanced level. 
     
    The differences between idioms and expressions in terms of literal and semantic meaning not only pose challenges for language learners, but also raise questions for cloze test models, where it should be taken into consideration how to model the meaning of idioms and expressions differently in order to achieve better performances. Many baseline models currently available for cloze test use random initialization when applying BERT to learn representation of the context and the idiom embedding.  But idioms and fixed expressions are different such that the literal meaning of the phrases may or may not be consistent with their semantic or contextual meaning. 
    
    In light of the above, we aim to achieve the following contributions for cloze test corpus construction and baseline model studies: 
    
    (1) We propose InDEX, an Indonesian Indiom and Expression dataset for cloze test.  The dataset contains 10438 unique sentences for 289 idioms and expressions for which we generate 15 different types of distractors, resulting in 100K cloze-style questions. 
It is the first dataset available for training Indonesian cloze test reading comprehsion models for both idioms and fixed expressions
   (2) We explore different ways to combine static and contextual representations for idioms and fixed expressions by distinguishing literal and semantic meaning representation. We draw conclusions on how best to initialize embedding for idioms and fixed expressions differently to achieve the best baseline performance.  Our dataset and models are freely available upon request.
\bgroup
\def\arraystretch{1.3} 
 \begin{table*}[h]
\begin{center}
\begin{tabular}{c|c|c|c}
\hline \hline
\bf Phrase Types & \bf Indonesian & \bf Literal Meaning & \bf Semantic Meaning \\ \hline
 & ada ekor & there a tail & tracking down sb. or sth. by following clues \\
Idioms & setengah hati &	half heart	& double-minded\\
& panjang lebar &	width length &	from beginning to end\\
\hline
& gaya bahasa	& style language	&style of writing\\
Fixed Expressions & padang pasir &	field sand	& desert\\
& juru bicara	& master talk&	spokesman\\
\hline \hline
\end{tabular}
\end{center}
\caption{\label{table1} Comparing Literal and Semantic Meaning of Indonesian Idioms and Fixed Expressions }
\end{table*}
\egroup

\section{Related Research}
\label{Related}

\subsection{Idioms and Fixed Expressions}

	Previous researchers used annotation recognition, definition replacement and other methods to better distinguish idioms from fixed expressions. Hashimoto \cite{b1} used disambiguation knowledge and grammatical dependency tree to build idiom recognizer and distinguish phrases into literal meaning and habitual meaning. Cook\cite{b2} used unsupervised methods to classify potential phrases into idioms and fixed expressions, and achieved performance similar to that of supervised methods. Muzny \cite{b7} matched phrases with their definitions through word sense disambiguation algorithm, which greatly improved idiom detection performance. Liu \cite{b8} proposed a supervised integration model to classify idioms and literal expressions, using late and early fusion methods. Most recent work on Indonesian idioms and expressions is comparative analysis from a linguistic research perspective against idioms and expressions from other languages (\cite{b9,b10,b11}). The study of idioms in non-common languages such as Indonesian is much rarer than that in languages such as English, and therefore desires more attention.

\subsection{Cloze test datasets and models}

 Researchers have proposed many challenging cloze task data sets including: CNN/daily mail\cite{b12}, children's book test (CBT) \cite{b13}, ClicR \cite{b14} in the medical field, Who-did-what \cite{b15} from LDC English Gigaword Newswire, and Cloth \cite{b16} from the training questions of Chinese high school and college entrance exams.

In study of cloze-test for idioms, Hashimoto \cite{b1} proposed a Japanese idiom corpus with 102846 sentences for 146 vague idioms. Yu Shiwen et al. \cite{b5} constructed a Chinese idiom knowledge base (CIKB). Zheng et al. \cite{b6} published the first large-scale cloze Chinese idiom data set CHID for machine reading comprehension, which contains more than 50K paragraphs and 60K blanks.

We observe from these previous study that the current cloze test baseline mainly uses random initialization for phrase embedding and capture the meaning of idioms from the given context. However, idioms and fixed expressions are different such that the former's semantic meaning can not be derived from their literal meaning, but the latter can. Therefore we want to model this discrepancy for a stronger baseline for idioms and fixed expressions as well. 

\section{Dataset Construction}

   Cloze test generally involves construction of the stem (i.e. a sentence with the target word removed), the key (i.e. the target word), and the set of distractors (i.e. plausible but erroneous words for completing the sentence). Table~\ref{table2} provides an example of cloze test question for Indonesian. 
 \begingroup
\renewcommand{\arraystretch}{1.3}
\begin{table*}[h]
\begin{center}
\begin{tabular}{l|l|l|l}
  \hline \hline
  \multirow{4}{*}{\textbf{Stem:}} & \multicolumn{3}{p{11cm}}{Akibat \_\_\_\_\_\_  dari Sangkuni, Pandu pun terlibat dalam perang melawan muridnya sendiri, yaitu seorang raja raksasa dari negeri Pringgadani bernama Prabu Tremboko.}
   \tabularnewline
   &\multicolumn{3}{p{11cm}}{[\textit{English}] As a result of Sangkuni's \_\_\_\_\_\_ Pandu was also involved in the war against his own disciples, a giant king from Pringgadani country named Prabu Tremboko.}\\
   \hline
\multirow{4}{*}{\textbf{Candidates:}} & \textbf{Key:} & adu domba & [\textit{English}] alienate one person from another\tabularnewline
\cline{2-4}
&\multirow{3}{*}{\textbf{Distractors:}} & empat mata	& [\textit{English}] whispering\tabularnewline
& & naik daun &	[\textit{English}] opportunity knocks \tabularnewline
& & air batu &	[\textit{English}] sorbet \tabularnewline
\hline \hline
\end{tabular}
\end{center}
\caption{\label{table2} Example Cloze Test Questions for Indonesian Idioms }
\end{table*}
\endgroup

To construct a high-quality cloze test dataset for idioms and fixed expressions, we need the following: phrases, sentences as stems, and distractors. We present our procedures as follows.
  
\subsection{Phrase Extraction}
\label{ssec:phrase}

Our idioms and fixed expressions are extracted from one of the most authoritative dictionary of Indonesian idioms and expressions written by J.S. Badudu in 1978 titled "Kamus Ungkapan Bahasa Indonesia". We manually extracted a total of 919 phrases, their illustrative sentences, and their definitions, all in Indonesian.  We invited two Associate Professors and 25 senior students in Indonesian Department to verify the extracted phrases, sentences, and definitions. They manually classify the phrases into two groups of idioms and fixed expressions. We use these two types of phrases and their definitions in Indonesian in the following dataset preparation and baseline model design.

\subsection{Stem Selection}
\label{ssec:stem}
To prepare sentences for the cloze test questions, we use a corpus of 8.74 million Indonesian Wikipedia articles collected in-house. We extracted 160K sentences containing the 919 phrases from the dictionary. We observe that the number of sentences available for each phrase differ drastically ranging from zero to more than 50K.  In order to ensure a properly-balanced sentence distribution by phrases for model training purposes, we only keep phrases with the number of available sentences ranging between 20 and 40.  Therefore, our final list of 289 phrases include 42 idioms and 247 fixed expressions. We further filter out sentences with the following heuristics:

\begin{itemize}
\item Sentences with less then ten words; 
\item Sentences containing characters from other languages ;
\item Sentences with the number of punctuations exceeding one-third of the sentence length;
\item Sentences ending with ":" (colon); 
\item Sentences identified as sequences of segments instead of full sentence;
\end{itemize}
\subsection{Cloze Test Question Generation}

Distractor generation is important in MC cloze test preparation. A good distractor must maintain the grammatical characteristics of the key, and similar in semantics with the key as well \cite{b20}. We design three strategies to generate five types of distractors.\\
\textbf{Random selection}. We select three distracters randomly for each question from the vocabulary list. \\
\textbf{Homonymy}. We calculate the Levenshtein Distance between the words in the vocabulary, and select the three closest neighbors to the key as distractors.\\
\textbf{Semantic similar words}. We use Word2Vec \cite{b21,b22} to produce word embeddings for the words, and calculate their pairwise cosine similarities. For each key, we select its top three neighbors in similarity as distractors. \\
To construct candidates for each stem, we use the three strategies to generate 5 types of candidate sets.
\begin{itemize}
\item Random distractors (\textbf{3RD}): all three distractors are randomly selected; 
\item Homonymy distractors (\textbf{3HD}): all three distractors are homonymies; 
\item Semantic Similar Distractors (\textbf{3SD}): all three distractors are semantically similar words; 
\item \textbf{1H2S}: 1 homonymy distractor and 2 semantic similar distractors; 
\item \textbf{1S2H}: 1 semantic similar distractor and 2 homonymy distractors. 
\end{itemize}
In addition to designing questions for idioms and fixed expressions separately where the candidates are from their own type, we also combine the phrases and all the sentences together and generate 5 types of distractors where the candidates include both idioms and fixed expressions. This design of questions will support our study on the different representation methods for idioms and fixed expressions. Table~\ref{data} presents the statistics of the InDEX corpus.
\begingroup
\renewcommand{\arraystretch}{1.3}
\begin{table*}[h]
\begin{center}
\begin{tabular}{ccccccc}
\hline \hline
\multirow{2}{*}{\textbf{Phrase Types}} & \multicolumn{5}{c}{\textbf{Types of Candidate Sets}}
 & \multirow{2}{*}{\textbf{Total Questions}} \\ \cline{2-6}
&  \bf 1S2H & \bf 1H2S & \bf 3RD & \bf 3SD & \bf 3H &  \\ \hline
Idioms&	1448	&1448&	1448	&1448	&1448&	7240\\
Fixed Expressions	&8990&	8990&	8990	&8990&	8990	&44950\\
Combined&	10438&	10438&	10438&	10438	&10438&	52190\\
\bf Total Questions	&20876	&20876&	20876&	20876	&20876	&104380\\
\hline \hline
\end{tabular}
\end{center}
\caption{\label{data} Idioms and Fixed Expression Cloze Test Data Statistics. }
\end{table*}
\endgroup

\section{Cloze Test Model Design}
\label{ssec:result}
   Idioms and fixed expressions are different such that idioms' meaning can not be deduced from the literal meanings of the words.  However, fixed expression's meaning is in general transparent and consistent with its literal meaning. Previous study on the cloze test model generally use BERT with random initialization of idiom embedding. Random initialized representation of phrases does carry any literal or semantic meaning of the phrase as input into the BERT-based reading comprehension model. Therefore for a cloze test dataset with for both idioms and fixed expression, we assume such a baseline does not consider the lexical and morphological differences between idioms and fixed expression, and fall short as baseline.
   
    We propose instead to investigate the different representations of idioms and fixed expressions as input for BERT-based model to achieve a well-defined baseline.  Word representation methods can be categorized into two classes: static word embedding based method, and contextual representation based method. Static word embeddings of different languages are pretrained in large monolingual corpora independently. Contextual representations can be obtained through multilingual pretraining, which encodes whole sentence as context. Combining the contextual representations with the word embeddings have been proved successful in machine translation. We propose to experiment with different embedding combination strategies for idioms only, fixed expression only, and mixture of both types of phrases for cloze test models.  
    
We divide the dataset into training, validation, and test sets for idioms, fixed expressions, and mixture model experiments as described in Table~\ref{stas}. 
\begingroup
\renewcommand{\arraystretch}{1.3}
\begin{table}[h]
\begin{center}
\begin{tabular}{ccccc}
\hline \hline
\bf Experiments & \bf Training & \bf Validation & \bf Test & \bf Total\\
 \hline
Idioms	& 5068 & 	1448 &	724	&7240\\
Fixed Expressions &	31465&	8990	&4495	&44950\\
Combined	&36533&	10438&	5219	&52190\\
\hline \hline
\end{tabular}
\end{center}
\caption{\label{stas} Experiment Data Statistics. }
\end{table}
\endgroup

\subsection{Test of Static Representation for Idioms and Fixed Expressions}

We first investigate and compare the static embedding for idiom and fixed expression separately using a BERT baseline model as described in Figure 2. We use Multilingual BERT \cite{b23} and Ind-BERT\cite{b24}. Multilingual BERT is pretrained in the same way as monolingual BERT while Ind-BERT is pretrained with Indonesian.
\begin{figure}
\begin{center}
\includegraphics[width=8cm]{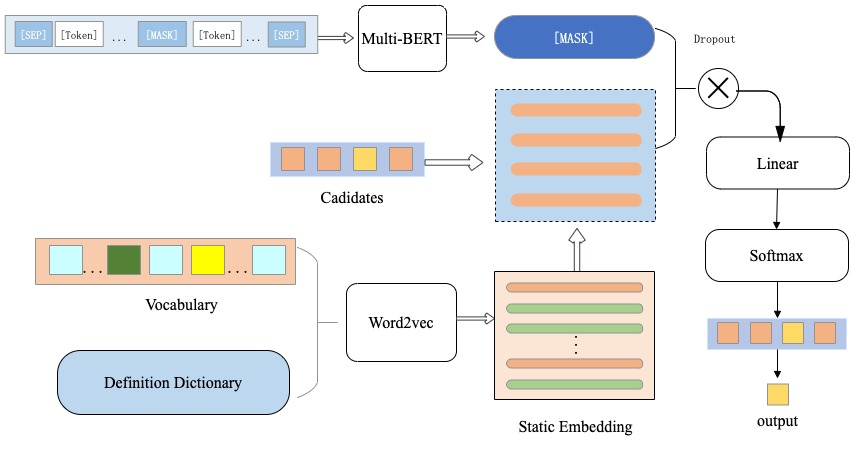}\\
\caption{\label{fig2} Cloze Test Model with Static Representations }
\end{center}
\end{figure}

We input the sentence with the key masked into the BERT model and extract the last hidden layer representation of the blank ($h_b$) and use it to score the candidates:

\begin{flalign}
h_b &= BERT(masked\_stem) \\ 
\alpha_i &= softmax(W(h_b^TE_i)+b
\end{flalign}

where $E_i$ is the embedding of each candidate term with three different implementations as discussed below, $W$ and $b$ are parameters to be learned, and the candidate with the highest $\alpha_i$ is chosen as the output answer. The hyper parameters used for training and testing the models are provided in Table~\ref{param}. Accuracy is calculated based on predicting correctly all candidates including the distractors to evaluate the neural language model's performance in general.

We implemented 3 different types of $E_i$ as static representations for idioms and fixed expressions as follows. $Idiom\_E_i^{Def}$  and $FixedExp\_E_i^{Def}$ are static representations using Skip-Gram, average word embedding, and phrase definition, where $w_k^{Def}$ refers to the $k (k \in[1,n])$ word in the definition of the phrase. $FixedExp\_E_j^{Lit}$ refers to the static representation by averaging the word embeddings learned with Skip-Gram of words composing the fixed expressions. We do not design a similar literal static embedding for idioms because literal representation does not contain the semantic meaning of idioms.

\begin{flalign}
Idiom\_E_i^{Def} &=\frac{1}{n}\sum_{k=1}^{n} word2vec(w_k^{Def})\\
FixedExp\_E_i^{Def} &= \frac{1}{n}\sum_{k=1}^{n} word2vec(w_k^{Def})\\
FixedExp\_E_j^{Lit} &= \frac{1}{m}\sum_{l=1}^{m} word2vec(w_l^{Lit})
\end{flalign}

\begingroup
\renewcommand{\arraystretch}{1.2}
\begin{table}[h]
\begin{center}
\begin{tabular}{cc}
\hline \hline
\bf Hyperparameters &	\bf Values\\
 \hline
 Optimizer	& AdamW\\
Criterion	& Cross Entropy Loss\\
Batch Size	& 16\\
Max Sequence Length	& 128\\
Learning Rate	& 2e-5\\
Epochs&	50\\
Dropout&	0.1\\
KFold	& 10\\
\hline \hline
\end{tabular}
\end{center}
\caption{\label{param} Hyperparameters }
\end{table}
\endgroup

\subsection{Static and Contextual Representations for Mixture of Idioms and Fixed Expressions Cloze Test}
\label{combine}

Our second test is to explore the combination strategies of static and contextual embeddings for cloze test with both idioms and fixed expression questions. The mixture of two types phrases with different semantic characteristics may pose more challenges to language learners. It is quite plausible to have a mixture of idioms and expressions in reading comprehension tests.  This type of mixture will also bring more challenges to BERT-based model as well.

\begin{figure*}[h]
\begin{center}
\includegraphics[width=11cm]{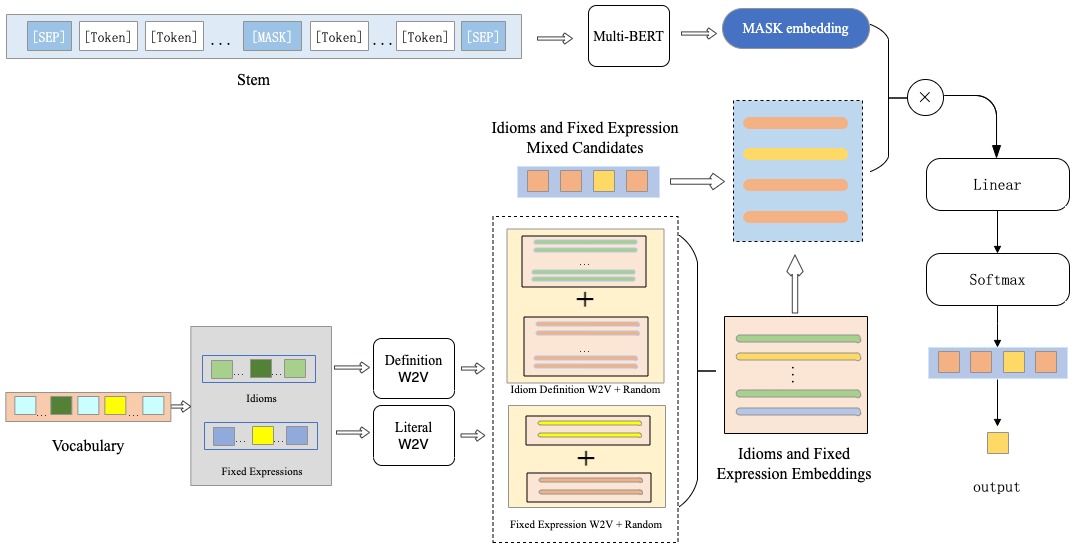}\\
\caption{\label{fig3} Mixture Cloze Test Model for Idioms and Fixed Expression Combined }
\end{center}
\end{figure*}

   The model structure is as presented in Figure~\ref{fig3}. We use the same model definitions as equations (1) and (2), the same static embedding implementation as equations (3) (4) and (5), and the same hyper parameters as in Table~\ref{param}. The differences are in the way that the different static representations are used for idioms and fixed expressions, and the combination among them as well. More details will be provided in the next section.

\section{Results and Analysis}
\label{sec:results}

We first present test of idioms as in Table~\ref{stat-context} and of fixed expression as in Table~\ref{stat-context-fix}. We may draw observations as follows:

\begingroup
\renewcommand{\arraystretch}{1.3}
\begin{table}[h]
\begin{center}
\begin{tabular}{cccccc}
\hline \hline
\multirow{2}{*}{\textbf{Model}} &	\multicolumn{5}{c}{\textbf{Types of Candidate Sets}} \\
\cline{2-6}
& \bf 3RD&	\bf 3SD&	\bf 3HD&	\bf 1H2S&	\bf 1S2H\\
\hline
$Random$ (Multi-BERT) &	0.792&	0.698&	0.704&	0.748&	0.767\\
$Idiom\_E_i^{Def}$ (Multi-BERT)	& \bf 0.799	& \bf0.711&	\bf0.723&	\bf0.761&	\bf0.786\\
$Random$ (Ind-BERT) &	0.862&\bf	0.811	& \bf 0.849	&0.818 & \b	0.849\\
$Idiom\_E_i^{Def}$(Ind-BERT)	&\bf 0.874&	0.805&	 0.836&	\bf0.824	& 0.836\\
 \hline \hline
\end{tabular}
\end{center}
\caption{\label{stat-context} Combining Static and Contextual Representation for Idioms }
\end{table}
\endgroup

\begingroup
\renewcommand{\arraystretch}{1.3}
\begin{table*}[h!]
\begin{center}
\begin{tabular}{cccccc}
\hline \hline
\multirow{2}{*}{\textbf{Model}} &	\multicolumn{5}{c}{\textbf{Types of Candidate Sets}} \\
\cline{2-6}
& \bf 3RD&	\bf 3SD&	\bf 3HD&	\bf 1H2S&	\bf 1S2H\\
\hline
$Random$ (Multi-BERT) & \bf0.832& \bf	0.783	& \bf 0.811	& 0.765	& \bf 0.806\\
$FixedExp\_E_i^{Def}$ (Multi-BERT)	& 0.801&0.771&0.770&	0.746&	0.757\\
$FixedExp\_E_j^{Lit}$ (Multi-BERT) & 0.808	&0.761	 & 0.801	& \bf 0.772	&0.792\\
\cline{2-6}
$Random$ (Ind-BERT) & 0.864	& 0.810&	0.845	&0.817	 & \bf0.854\\
$FixedExp\_E_i^{Def}$ (Ind-BERT) & 0.860	&\bf 0.827&\bf	0.846&	0.817	& 0.837\\
$FixedExp\_E_j^{Lit}$ (Ind-BERT) & \bf 0.865	& 0.822&	0.822& \bf	0.831&	0.839\\
 \hline \hline
\end{tabular}
\end{center}
\caption{\label{stat-context-fix} Combining Static and Contextual Representation for Fixed Expressions }
\end{table*}
\endgroup

(1) \textbf{The effect of distractor generation strategies with random initialized embedding}. For both idioms and fixed expressions, the 5 types of distractor demon-strate similar patterns of challenge for BERT model. In terms of prediction accuracy from low to high, for Multi-BERT, the order is 3SD $<$ 3HD $<$ 1H2S $<$ 1S2H $<$ 3RD.  For Ind-BERT, the accuracy from low to high is 3SD $<$ 1H2S $<$ 1S2H $=$ 3HD $<$ 3RD. In particular, Ind-BERT outperformed Multi-BERT in its prediction with homonymy distractors.

(2) \textbf{The effect of different static representations for idioms}. We observe from Table 6 that static embeddings learned with idiom definition outperform random initialized embedding for Multi-BERT, but lose 2-3 to random embedding for Ind-BERT. This partially confirms our hypothesis that definition embedding will provide additional semantic input for contextual representation. Furthermore, Ind-BERT outperformed Multi-BERT.

(3) \textbf {The effect of different static representations for fixed expressions}. We observe from Table~\ref{stat-context-fix} that embeddings based on expression definitions are not contributing to the Multi-BERT model, and fall short on Ind-BERT as well. Random initialization works for Multi-BERT, and lost to embeddings based on literal meaning for Ind-BERT. We may conclude that definition may not be helpful for fixed expression whose semantic meaning does not deviate much from definition. We may consider random initialization and literal static embeddings in the combination cloze test.

(4)\textbf{The effect of static representations for mixture of idioms and expressions}.  We now present results for different combinations of static embeddings with contextual embeddings for cloze test of idioms and fixed expression combined. Drawing from the conclusions from Table~\ref{stat-context} and Table~\ref{stat-context-fix}, we explore $Idiom\_E_i^{Def}$  and $FixedExp\_E_j^{Lit}$  as compared with random initializations.  As shown in Table~\ref{stat-context-mix},  for Multi-BERT, the best performances are given by  $Idiom\_E_i^{Def}+Random$ for Idioms and $Random$ for fixed expression.  The same observations can be drawn for Ind-BERT, where the combination of $Idiom\_E_i^{Def}+Random$ for Idioms and $Random$ for fixed expression works the best, and even better than for Multi-BERT.

\begingroup
\renewcommand{\arraystretch}{1.3}
\begin{table*}[h!]
\tiny
\begin{center}
\resizebox{\textwidth}{!}{%
\begin{tabular}{cp{2cm}p{2cm}ccccc}
\hline \hline
\bf Model &\bf Idioms &\bf Fixed\par \bf Expression & \bf3RD&	\bf3SD&	\bf3HD&	\bf1H2S&	\bf1S2H\\
\hline
\multirow{6}{*}{\bf Multi-\strut BERT} & $Random$ & $Random$ & 0.809&	0.776&	0.789&	0.774&	0.800\\
\cline{2-8}
&$Idiom\_E_i^{Def}$& $FixedExp\_E_j^{Lit}$ & 0.801&	0.750&	0.785&	0.760&	0.784\\
\cline{2-8}
& $Idiom\_E_i^{Def}$ & $Random$ &0.812&	0.750&	0.781&	0.765&	0.776\\
\cline{2-8}
& $Idiom\_E_i^{Def}$\par $+Random$& $FixedExp\_E_j^{Lit}$ &0.812&	0.750&	0.780&	0.765&	0.776\\
\cline{2-8}
&$Idiom\_E_i^{Def}$\par $+Random$ & $FixedExp\_E_j^{Lit}$\par$+Random$ &0.824&	0.848&	0.848&	0.820&	\bf0.844\\
\cline{2-8}
&$Idiom\_E_i^{Def}$\par $+Random$ & $Random$ &\bf0.851&\bf	0.853&\bf	0.854&	\bf0.830&	0.833\\
\hline

\multirow{6}{*}{\bf Ind-\strut BERT} & $Random$ & $Random$ &0.869&	0.825&	0.840&	0.817&	0.833\\
\cline{2-8}
& $Random$ & $FixedExp\_E_j^{Lit}$ &0.872&	0.820&	0.844&	0.821&	0.835\\
\cline{2-8}
& $Idiom\_E_i^{Def}$ & $Random$ &0.866&	0.829&	0.839&	0.814&	0.839\\
\cline{2-8}
& $Idiom\_E_i^{Def}$ & $FixedExp\_E_j^{Lit}$ &0.866&	0.829&	0.840&	0.814&	0.839\\
\cline{2-8}
& $Idiom\_E_i^{Def}$\par $+Random$ & $FixedExp\_E_j^{Lit}$\par$+Random$ &0.874&	0.880&	\bf0.886&	0.863&	0.883\\
\cline{2-8}
& $Idiom\_E_i^{Def}$\par $+Random$ & $Random$ &\bf0.879&\bf	0.883&	0.883&	\bf0.875&	\bf0.887\\
\hline \hline
\end{tabular}}
\end{center}
\caption{\label{stat-context-mix} Test of Combining Static and Contextual Representation for Mixture Cloze Test of Idioms and Fixed Expressions }
\end{table*}
\endgroup

\section{Conclusions}
This paper aims to address the following three research gaps. First is the lack of the idiom cloze test corpus and baseline model for low-resourced languages such as Indonesian. The second is that current study on cloze test questions generally focus on idioms alone, without consideration of and comparisions with fixed expressions. The third is that current baseline models for cloze test of idioms generally applies random initialization for learning contextual representation of idioms with BERT. Because of the differences between idioms and fixed expressions in terms of literal meaning and semantic meaning, we propose instead to explore static and dynamic embedding and their combination for idiom-only, fixed expression-only, and mixture of idioms and fixed expression cloze test models. 

We contribute to cloze test study by building the first Indonesian Idioms and Fixed Expression cloze test corpus with one of the most authoritative dictionary and with careful design of distractors. We explore different static embeddings for idioms and expressions based on definitions and literal word compositions. Experimentations show that idiom definition combined with random initialization will better support cloze test of idioms especially in mixture model with fixed expressions. While for fixed expressions with no special meaning from its literal, random initialization is sufficient for separate or mixture cloze test. Future research may include more sophisticated distractor generation, and comparisons with human test results. 


\end{document}